\documentclass{article}

\usepackage{times}
\usepackage{graphicx} 
\usepackage{subfigure} 

\usepackage{natbib}

\usepackage{algorithm}
\usepackage{algorithmic}

\usepackage{amssymb}
\usepackage{booktabs}
\usepackage{amsmath}
\usepackage{etoolbox}
\usepackage{xspace}

\newtoggle{mt}
\togglefalse{mt}

\usepackage{hyperref}
\usepackage{breakurl}

\usepackage[accepted]{icml2017}

\icmltitlerunning{Revisiting Activation Regularization for Language RNNs}

\begin{document}

\newcommand{\ltwo}{$L_2$\xspace}

\twocolumn[
\icmltitle{Revisiting Activation Regularization for Language RNNs}



\icmlsetsymbol{equal}{*}

\begin{icmlauthorlist}
\icmlauthor{Stephen Merity}{sfr}
\icmlauthor{Bryan McCann}{sfr}
\icmlauthor{Richard Socher}{sfr}
\end{icmlauthorlist}
\icmlaffiliation{sfr}{Salesforce Research, Palo Alto, USA}
\icmlcorrespondingauthor{Stephen Merity}{smerity@salesforce.com}

\icmlkeywords{machine learning, deep learning, RNN, activation regularization, ICML, LSTM}

\vskip 0.3in
]



\printAffiliationsAndNotice{}  

\begin{abstract} 
Recurrent neural networks (RNNs) serve as a fundamental building block for many sequence tasks across natural language processing.
Recent research has focused on recurrent dropout techniques or custom RNN cells in order to improve performance.
Both of these can require substantial modifications to the machine learning model or to the underlying RNN configurations.
We revisit traditional regularization techniques, specifically \ltwo regularization on RNN activations and slowness regularization over successive hidden states, to improve the performance of RNNs on the task of language modeling.
Both of these techniques require minimal modification to existing RNN architectures and result in performance improvements comparable or superior to more complicated regularization techniques or custom cell architectures.
These regularization techniques can be used without any modification on optimized LSTM implementations such as the NVIDIA cuDNN LSTM.
\end{abstract}

\section{Introduction}

The need for effective regularization methods for RNNs has seen extensive focus in recent years.
While application of dropout \citep{Srivastava2014DropoutAS} to the input and output of an RNN has been shown to be effective \citep{Zaremba2014}, dropout is destructive when naively applied to the recurrent connections of an RNN.
When naive dropout is applied to the recurrent connections, it is almost impossible to retain information over long periods of time.

Given this fundamental issue, substantial work has gone into understanding and improving dropout when applied to recurrent connections.
Of these techniques, which we shall broadly refer to as recurrent dropout, some specific variations have gained popular usage.

Variational RNNs \citep{Gal2016ATG} drop the same network units at each timestep, as opposed to dropping different network units at each timestep.
By performing dropout on the same units at each timestep, destructive loss of the RNN hidden state is avoided and the same information is masked at each timestep.

Rather than dropping units, another tactic is to drop updates to given network units.
\citet{Semeniuta2016RecurrentDW} perform dropout on the input gate of the LSTM \citep{Hochreiter1997LongSM} but allow the forget gate to discard portions of the existing hidden state.
Zoneout \citep{Krueger2016} prevents hidden state updates from occurring by setting a randomly selected subset of network unit activations in $h_{t+1}$ to be equal to the previous activations from $h_t$.
Both of these act to prevent updates to the hidden state while preserving existing content.

On an extreme end, work has also been done to restrict the recurrent matrices in an RNN in order to limit their computational capacity.
Some RNN architectures only allow element-wise interactions \citep{Balduzzi2016,Bradbury2016,Seo2016}, removing the recurrent matrix entirely, while others act to restrict the capacity by parameterizing the recurrent matrix \citep{Arjovsky2016, Wisdom2016, Jing2016}.

Other forms of regularization explicitly act upon activations such as such as batch normalization~\citep{Ioffe2015BatchNA}, recurrent batch normalization~\citep{Cooijmans2016RecurrentBN}, and layer normalization~\citep{Ba2016LayerN}.
These all introduce additional training parameters and can complicate the training process while increasing the sensitivity of the model.
Norm stabilization~\citep{Krueger2015RegularizingRB} penalizes the model when the norm of an RNN's hidden state changes substantially between timesteps, achieving strong results in character language modeling on and phoneme recognition.

In this work, we revisit \ltwo regularization in the form of activation regularization (AR) and temporal activation regularization (TAR).
When applied to modern baselines that do not contain recurrent dropout or normalization techniques, AR and TAR achieve comparable or superior results.

Compared to other invasive regularization techniques which may require modifications to the RNN cell itself or complex model changes, both AR and TAR require no substantial modifications to the RNN or model.
This enables AR and TAR to be applied to optimized RNN implementations such as the cuDNN LSTM which can be many times faster than na\"{i}ve but flexible LSTM implementations.

\section{Activation Regularization}

\subsection{\ltwo activation regularization (AR)}

While \ltwo regularization is traditionally used on the weights of machine learning models (\ltwo weight decay), it could also be used on the activations.
We define AR as
\[
\alpha \, L_2 (m \odot h_t) 
\]
where $m$ is the dropout mask used by later parts of the model, $L_2(\cdot) = \lVert \cdot \rVert_2$ (\ltwo norm), $h_t$ is the output of the RNN at timestep $t$, and $\alpha$ is a scaling coefficient.

When applied to the output of a dense layer, AR penalizes activations that are substantially away from 0, encouraging the activations to remain small.
While acting implicitly rather than explicitly, this has similarities to the various batch or layer normalization techniques.

The \ltwo penalty on the RNN activations can be applied to $h_t$ or to $m \odot h_t$ (the dropped output used in the rest of the model).
In our experiments, we found that applying AR to $m \odot h_t$ was more effective than applying it to neurons not updated during the current optimization step.

\subsection{Temporal activation regularization (TAR)}

Adding a prior that minimizes differences between states has been explored in the past.
This broad concept falls under the broad concept of slowness regularization \citep{Hinton1989, Foldiak1991, Luciw2012, Jonschkowski2015LearningSR,Wen2015} which attempts to minimize $L(f(x_t), f(x_{t+1}))$ where $L$ is a loss function describing the distance between $f(x_t)$ and $f(x_{t+1})$ and $f$ is an arbitrary mapping function.

Temporal activation regularization (TAR) is a direct descendant of this slowness regularization, minimizing
\[
\beta \, L_2 (h_t - h_{t+1})
\]
where $L_2(\cdot) = \lVert \cdot \rVert_2$ (\ltwo norm), $h_t$ is the output of the RNN at timestep $t$, and $\beta$ is a scaling coefficient.

TAR penalizes any large changes in hidden state between timesteps, encouraging the model to keep the output as consistent as possible.
For the LSTM, the hidden state which is regularized is only $h_t$, not the long term memory $c_t$, though this could optionally be regularized in a similar manner.

\section{Experiments}

\subsection{Language Modeling}

We benchmark activation regularization (AR) and temporal activation regularization (TAR) applied to a strong non-variational LSTM baseline\footnote{PyTorch Word Level Language Modeling example: \url{https://github.com/pytorch/examples/tree/master/word_language_model}}.
The experiment uses a preprocessed version of the Penn Treebank (PTB) \citep{Mikolov2010} and WikiText-2 (WT2) \citep{Merity2016}.
All hyperparameters, including $\alpha$ for AR and $\beta$ for TAR, are optimized over the validation dataset.
The best found hyperparameters as determined by the validation results are then run on the test set.

\begin{table}
\center
\begin{tabular}{l|cc}
\toprule
\bf Model & \bf Parameters & \bf Validation \\
\midrule
$\alpha=0$ & 13M & $78.4$ \\
$\alpha=1$ & 13M & $76.2$ \\
$\alpha=3$ & 13M & $73.9$ \\
$\alpha=5$ & 13M & $73.7$ \\
$\alpha=7$ & 13M & $73.0$ \\
$\alpha=9$ & 13M & $74.0$ \\
\bottomrule
\end{tabular}
\caption{Results over the Penn Treebank for testing $\alpha$ coefficients for AR with base model $h=650, \beta=0, \text{dp}=0.5, \text{dp}_h=0.5$.}
\label{table:alphaonly}
\end{table}

\begin{table}
\center
\begin{tabular}{l|cc}
\toprule
\bf Model & \bf Parameters & \bf Validation \\
\midrule
$\beta=0$ & 13M & $78.4$ \\
$\beta=1$ & 13M & $77.2$ \\
$\beta=3$ & 13M & $75.2$ \\
$\beta=5$ & 13M & $74.4$ \\
$\beta=7$ & 13M & $74.1$ \\
$\beta=9$ & 13M & $74.7$ \\
\bottomrule
\end{tabular}
\caption{Results over the Penn Treebank for testing $\beta$ coefficients for TAR with base model $h=650, \alpha=0, \text{dp}=0.5, \text{dp}_h=0.5$.}
\label{table:betaonly}
\end{table}

\begin{table*}
\center
\begin{tabular}{l|ccc}
\toprule
\bf Model & \bf Parameters & \bf Validation &  \bf Test \\
\midrule
PTB, LSTM (tied) $h=650, \text{dp}=0.5, \text{dp}_h=0.4$ & 13M & $78.2$ & $74.8$ \\
PTB, LSTM (tied) $h=950, \text{dp}=0.6, \text{dp}_h=0.5$ & 24M & $75.3$ & $72.2$ \\
PTB, LSTM (tied) $h=1500, \text{dp}=0.75, \text{dp}_h=0.5$ & 51M & $71.3$ & $68.3$ \\
\midrule
PTB, LSTM (tied) $h=650, \alpha=5, \beta=2, \text{dp}=0.5, \text{dp}_h=0.4$ & 13M & $72.0$ & $68.9$ \\
PTB, LSTM (tied) $h=950, \alpha=6, \beta=4, \text{dp}=0.6, \text{dp}_h=0.5$ & 24M & $70.2$ & $66.9$ \\
PTB, LSTM (tied) $h=1500, \alpha=4, \beta=4, \text{dp}=0.75, \text{dp}_h=0.5$ & 51M & $68.2$ & $65.4$ \\
\bottomrule
\end{tabular}
\caption{Single model perplexity results over the Penn Treebank.
Models noting \textit{tied} use weight tying on the embedding and softmax weights.
The top section contain models without AR or TAR with the bottom section containing equivalent models using them.
}
\label{table:SelfPTBresults}
\end{table*}

\begin{table*}
\center
\begin{tabular}{l|ccc}
\toprule
\bf Model & \bf Parameters & \bf Validation &  \bf Test \\
\midrule
\citet{Inan2016} - Variational LSTM  (tied) ($h=650$) & 28M & $92.3$ & $87.7$ \\
\citet{Inan2016} - Variational LSTM  (tied) ($h=650$) + augmented loss & 28M & $91.5$ & $87.0$ \\
\midrule
WT2, LSTM (tied) $h=650, \text{dp}=0.5, \text{dp}_h=0.4$ & 28M & $88.8$ & $84.9$ \\
\midrule
WT2, LSTM (tied) $h=650, \alpha=5, \beta=2, \text{dp}=0.5, \text{dp}_h=0.4$ & 28M & $85.8$ & $81.8$ \\
\bottomrule
\end{tabular}
\caption{Results over WikiText-2.
The increases in parameters compared to the models on PTB are due to the larger vocabulary.
Models noting \textit{tied} use weight tying on the embedding and softmax weights.
}
\label{table:SelfWT2results}
\end{table*}

\textbf{PTB:} As the Penn Treebank is a small dataset, preventing overfitting is of considerable importance and a major focus of research.
Almost all competitive models rely upon a form of recurrent dropout to ensure the RNN does not overfit through drastic changes in the hidden state.
Other aggressive dropout techniques, such as performing dropout on the embedding layer such that entire words are dropped from a sequence, are also frequently used.

\textbf{WT2:} WikiText-2 is a dataset approximately twice as large as PTB but with a vocabulary three times larger.
The text is also tokenized and processed in a manner similar to datasets used for machine translation using the Moses tokenizer \citep{Koehn2007MosesOS}.
\iftoggle{mt}{We use this dataset as an intermediate step to test potential trends as we transfer these results to machine translation.}

\textbf{Experiment details:}
All experiments use a model containing a two layer RNN.
The AR and TAR loss are only applied to the output of the final RNN layer, not to all layers.
For the majority of experiments, we follow the medium model size of \citet{Zaremba2014}: a two layer RNN with 650 hidden units in each layer.

For training the model, stochastic gradient descent (SGD) without momentum was used for up to 80 epochs.
The learning rate began at $20$ and was divided by four each time validation perplexity failed to improve.
\ltwo weight regularization of $10^{-7}$ was used over all weights in the model and gradients with norm over 10 were rescaled.
Batches consist of 20 examples with each example containing 35 timesteps.
The loss was averaged over all examples and timesteps.
All embedding weights were uniformly initialized in the interval $[-0.1, 0.1]$ and all other weights were initialized between $[-\frac{1}{\sqrt{H}}, \frac{1}{\sqrt{H}}]$, where $H$ is the hidden size.

For dropout, we have two different parameters, $\text{dp}$ and $\text{dp}_h$.
$\text{dp}$ is the dropout rate used on the word vectors and the final RNN output.
$\text{dp}_h$ is the dropout rate used on the connection between RNN layers.
All models use weight tying between the embedding and softmax layer \citep{Inan2016, Press2016}.

\textbf{Evaluating AR and TAR independently on PTB:}
To understand the potential of AR and TAR, we investigate their impact on language model perplexity when used independently in Table \ref{table:alphaonly} (AR) and Table \ref{table:betaonly} (TAR).
While both result in a substantial reduction in perplexity, AR results in the strongest improvement of $5.3$, while TAR only achieves $4.3$.
The drops achieved by this are equivalent to using an LSTM model with twice as many parameters - a substantial improvement given the simplicity of AR and TAR.

\textbf{Evaluating AR and TAR jointly on PTB}:
When both AR and TAR are used together, we found the best result was achieved by decreasing $\alpha$ and $\beta$, likely as the model was over-regularized otherwise.
In Table \ref{table:SelfPTBresults} we present PTB results for three different model sizes comparing models without AR/TAR to those which use both.
The model sizes $h \in [650, 950, 1500]$ were chosen to be comparable in size to other published results.
With both AR and TAR, the smallest model has an improvement of $6.2$ over the baseline model.
The improvements continue for the two larger size models, $h=950$ and $h=1500$, though the gains fall off as the model size is increased.

\begin{table*}
\center
\begin{tabular}{l|ccc}
\toprule
\bf Model & \bf Parameters & \bf Validation &  \bf Test \\
\midrule
\citet{Zaremba2014} - LSTM (medium) & 20M & $86.2$ & $82.7$ \\
\citet{Zaremba2014} - LSTM (large) & 66M & $82.2$ & $78.4$ \\
\citet{Gal2016ATG} - Variational LSTM (medium) & 20M & $81.9 \pm 0.2$ & $79.7 \pm 0.1$ \\
\citet{Gal2016ATG} - Variational LSTM (medium, MC) & 20M & $-$ & $78.6 \pm 0.1$ \\
\citet{Gal2016ATG} - Variational LSTM (large) & 66M & $77.9 \pm 0.3$ & $75.2 \pm 0.2$ \\
\citet{Gal2016ATG} - Variational LSTM (large, MC) & 66M & $-$ & $73.4 \pm 0.0$ \\
\citet{Kim2016} - CharCNN & 19M & $-$ & $78.9$ \\
\citet{Merity2016} - Pointer Sentinel-LSTM & 21M & $72.4$ & $70.9$ \\
\citet{Inan2016} - Variational LSTM (tied) + augmented loss & 24M & $75.7$ & $73.2$ \\
\citet{Inan2016} - Variational LSTM (tied) + augmented loss & 51M & $71.1$ & $68.5$ \\
\citet{Zilly2016} - Variational RHN (tied) & 23M & $67.9$ & $65.4$ \\
\citet{Zoph2016} - NAS Cell (tied) & 25M & $-$ & $64.0$ \\
\citet{Zoph2016} - NAS Cell (tied) & 54M & $-$ & $62.4$ \\
\midrule
PTB, LSTM (tied) $h=650, \alpha=5, \beta=2, \text{dp}=0.5, \text{dp}_h=0.4$ & 13M & $72.0$ & $68.9$ \\
PTB, LSTM (tied) $h=950, \alpha=6, \beta=4, \text{dp}=0.6, \text{dp}_h=0.5$ & 24M & $70.2$ & $66.9$ \\
PTB, LSTM (tied) $h=1500, \alpha=4, \beta=4, \text{dp}=0.75, \text{dp}_h=0.5$ & 51M & $68.2$ & $65.4$ \\
\bottomrule
\end{tabular}
\caption{
Single model perplexity on validation and test sets for the Penn Treebank language modeling task.
Models noting \textit{tied} use weight tying on the embedding and softmax weights.
}
\label{table:PTBresults}
\end{table*}

\textbf{Comparing to state-of-the-art PTB:}
In Table \ref{table:PTBresults} we summarize the current state of the art models in language modeling over the Penn Treebank.

The largest LSTM we train ($h=1500$) achieves comparable results to the Recurrent Highway Network (RHN) \cite{Zilly2016}, a human developed custom RNN architecture, but with approximately double the number of parameters.
Although the LSTM uses twice as many parameters, the RHN runs a cell 10 times per timestep (referred to as recurrence depth), resulting in far more computation.
This would likely result in the RHN being slower than the larger LSTM model during both training and prediction, especially when factoring in optimized LSTM implementations such as NVIDIA's cuDNN LSTM.

We also compare to the Neural Architecture Search (NAS) cell \citep{Zoph2016}.
While \citet{Zoph2016} do not report any of the hyperparameters or what type of dropout they used for their Penn Treebank result, they do note that they performed an extensive hyperparameter search over learning rate, weight initialization, dropout rates, and decay epoch in order to produce their best performing model.
It is possible that a large contributor to their improved result was in these tuned hyperparameters as they did not compare their NAS cell results to a standard or variational LSTM cell that was subjected to the same extensive hyperparameter search.
Our largest LSTM results are $3$ perplexity higher in comparison but have not undergone extensive hyperparameter search, do not use additional regularization techniques such as recurrent or embedding dropout, and do not use a custom RNN cell.

\textbf{WikiText-2 Results:}
We compare our WikiText-2 results to \citet{Inan2016} who introduced weight tying between the embedding and softmax weights.
While we did not perform any hyperparamter search over the coefficient values of $\alpha$ and $\beta$ for AR and TAR, instead using the best results from PTB, we find them to still be quite effective.
The baseline LSTM already achieves a $2.1$ perplexity improvement over the variational LSTM models from \citet{Inan2016}, including one which uses an augmented loss that modifies standard cross entropy with temperature and a KL divergence based loss.
When the AR and TAR parameters optimized over PTB are used, perplexity falls an additional $3.1$ perplexity.
This is not as strong an improvement as seen on the PTB dataset and may be due to the increased complexity of the dataset (larger vocabulary meaning a longer tail of usage, different genre, and so on) or may just be due to the lack of hyperparamter tuning.

\textbf{AR and TAR for GRU and $\tanh$ RNN:}
While neither the GRU \citep{Cho2014} or $\tanh$ RNN are traditionally used in language modeling, we wanted to see the generality of AR and TAR to other types of RNN cells.
We applied the best values of $\alpha$ and $\beta$ for an LSTM cell to the GRU and $\tanh$ RNN on PTB without any further search in Table \ref{table:RNNPTBresults}.
These values are likely quite suboptimal but are sufficient for illustrative purposes.
For the GRU, perplexity improved by $2.2$ from the baseline.
This is a positive sign given the impact of these regularization techniques on a GRU are quite different to that of an LSTM.
The LSTM only has $h_t$ subjected to AR and TAR, leaving the long term memory $c_t$ unregularized, but the GRU uses $h_t$ both as output at that timestep and as the hidden state input for the next timestep.
For the $\tanh$ RNN, the model did not train to acceptable levels at all without the application of AR and TAR.
For the $\tanh$ RNN, TAR likely forced the recurrent matrix to learn an identity function in order to ensure $h_t$ could produce $h_{t+1}$.
This would be important given the weights in this model were randomly initialized and suggests TAR acts as an implicit identity initialization constraint \citep{Le2015}.

\begin{table*}
\center
\begin{tabular}{l|ccc}
\toprule
\bf Model & \bf Parameters & \bf Validation &  \bf Test \\
\midrule
PTB, RNN (tied) $h=650, \text{dp}=0.5, \text{dp}_h=0.4$ & 13M & $712.3$ & $667.5$ \\
PTB, RNN (tied) $h=650, \alpha=5, \beta=2, \text{dp}=0.5, \text{dp}_h=0.4$ & 13M & $232.1$ & $227.8$ \\
\midrule
PTB, GRU (tied) $h=650, \text{dp}=0.5, \text{dp}_h=0.4$ & 13M & $86.1$ & $83.3$ \\
PTB, GRU (tied) $h=650, \alpha=5, \beta=2, \text{dp}=0.5, \text{dp}_h=0.4$ & 13M & $83.9$ & $81.1$ \\
\bottomrule
\end{tabular}
\caption{
Single model perplexity results over the Penn Treebank for $\tanh$ RNN and GRU. Neither cell are traditionally used for language modeling but this demonstrates the generality for AR ($\alpha$) and TAR ($\beta$).
Values for $\alpha, \beta$ taken from best LSTM model with no search.
Models noting \textit{tied} use weight tying on the embedding and softmax weights.
}
\label{table:RNNPTBresults}
\end{table*}

\section{Conclusion}

In this work, we revisit \ltwo regularization in the form of activation regularization (AR) and temporal activation regularization (TAR).
While simple to implement, activity regularization and temporal activity regularization are competitive with other far more complex regularization techniques and offer equivalent or better results.
The improvements that these techniques provide can likely be combined with other regularization techniques, such as the variational LSTM, and may lead to further improvements in performance as well, especially if subjected to an extensive hyperparameter search.




\section*{Sample generated text}

For generating text samples, words were sampled using the standard generation script contained in the PyTorch word level language modeling example.
WikiText-2 was used given the larger vocabulary and more realistic looking text.
Neither the $\langle eos \rangle$ token nor the $\langle unk \rangle$ were allowed to be selected.
Each paragraph is a separate sample of text with the tokens following Moses \citep{Koehn2007MosesOS}, joining words with @-@ and dot-decimal split to a @.@ token.

\rule{0.3\textwidth}{1pt}

" Something Borrowed " is the second episode of the fourth season of the American comedy television series The X @-@ Files . The episode was written by David McCarthy and directed by Mark Sacks . It aired in the United States on November 30 , 2011 , as a two @-@ episode episode, watched by 4 @.@ 9 million viewers and was the highest rated show on the Fox network .

The work of Olivier 's , a large 1950s table with the
center of a vinyl beam , was used for bony motifs from the upper @-@ production model via the Club van
X . The modified works were released in the museum , which gave its namesake to the visual designers in
Hong Kong .

The first prototype was released for the PlayStation 4 , containing the 2 @.@ 5 part series , with 3 @.@ 5 million copies sold . In October 2010 , Activision announced that both the game and the main gameplay was `` downloadable '' . The first game , titled Snow : The Game of the Battlefield 2 : The Ultimate Warrior , was the third anime game , and was released in August 2016 .

The German Land Forces had been reversed in the early 1990s , although the Soviet Union continued to deter NDH forces in the nation . The area was moved to Sarajevo , and the troops were despatched to the National Register of Historic Places in the summer of 1918 for the establishment of full political and social parties . The Polish language was protected by the Soviet Union , which was the first Polish continental conflict of the newly formed Union in North America , and the Polish Front with the last of the Polish Communist Party .

\bibliography{references}
\bibliographystyle{icml2017}

\end{document}